\documentclass{article}
\usepackage{isipta}

\newtheorem{Theorem}{Theorem}
\newtheorem{Definition}{Definition}
\newtheorem{Claim}{Claim}
\newtheorem{Lemma}{Lemma}

\title{What is the plausibility of probability?(revised 2003, 2015)}
\author{Stefan Arnborg \\
	Royal Institute of Technology KTH \\
	stefan@nada.kth.se\\
{\bf Gunnar Sj\"odin}\\
Swedish Institute of Computer Science\\
sjodin@sics.se
        }
\begin{document}

\maketitle

\begin{abstract}

We present and examine a result related to uncertainty reasoning, namely
that a certain plausibility space of Cox's type can be uniquely embedded in a
minimal ordered field. This, although a purely mathematical result, can be
claimed to imply that every rational method to reason with uncertainty
must be based on sets of extended probability distributions, where extended
probability is standard probability extended with infinitesimals.

This claim must be supported by some argumentation of non-mathematical type,
however, since pure mathematics does not tell us anything about the world.
We propose one such argumentation, and relate it to results from the literature
of uncertainty and statistics. 

In an added retrospective section we discuss some developments in the area 
regarding countable additivity, partially ordered domains and robustness, and philosophical
stances on the Cox/Jaynes approach since 2003. We also
show that the most general partially ordered plausibility calculus embeddable in a ring
can be represented as a set of extended probability distributions or, in algebraic terms,
is a subdirect sum of ordered fields. In other words, the robust Bayesian
approach is universal. 
This result is exemplified by relating Dempster-Shafer's 
evidence theory to robust Bayesian analysis.

\end{abstract}

\begin{keywords}
belief, plausibility, probability, Bayesianism
\end{keywords}

\section{Introduction}

We consider plausibility spaces of the type defined by Cox\cite{Cox46},
with auxiliary functions $F$, $S$ and $G$ used for computing
plausibilities of conjunctions, complements and disjunctions,
respectively. The domain of plausibility values with the auxiliary
functions and some ordering relation $\leq$ is called a
{\em plausibility space}\footnote{We use the term space instead of measure or algebra in order to avoid confusion with terminology in related work.}, where $a \leq b$
means that an event with plausibility value $a$ is equally or less plausible
than one with the value $b$. A system of conditional propositions or events
with plausibility values is a {\em plausibility model}
\footnote{Model is one of several possible terms; it suggests the analoguous use of models in logic.}. Standard probability
is one possible plausibility space, where the domain is the
real numbers in $[0,1]$ and the functions $F$ and $G$
are multiplication and addition, respectively.
Extended probability is standard probability extended with infinitesimal
probabilities. 

We will motivate a definition, prove two theorems and justify
a claim. 
{\bf Definition~1} below of proper ordered plausibility spaces and models
lists a number of properties we assume for the 
domain of plausibility values and the combination functions $F$ and $G$ used to 
find plausibilities of conjunctions and disjunctions when we have the plausibilities of their operands. These properties are similar to those assumed by Acz\'el and Cox. In particular,
$F$ and $G$ satisfy the algebraic laws of the operators $\times$ and $+$ of a ring and have certain monotonicity properties.

{\bf Theorem 1} says that a proper ordered plausibility space can be embedded in an ordered field
 where the field operators $\times$ and $+$ are
extensions of $F$ and $G$. An ordered field is a ring where certain monotonicity and solvability properties hold,
the best known examples being the fields of reals ${\bf R}$ and rationals ${\bf Q}$ . But there are more complex ordered fields that may be relevant and which we will discuss.

In order to motivate Definition~1, we investigate several assumptions, some old and some new, 
which state desirable properties of a plausibility space. We argue for the acceptance of a number of these assumptions, which we call
Sufficiency, Monotonicity, Propositional limit,
Refinability and Closedness.

{\bf Theorem~2} says that a plausibility space satisfying the above assumptions is a proper ordered plausibility space
as defined in Definition~1. Our claim is:

\begin{Claim}

Theorems 1 and 2 imply that the most general rational way to deal with
uncertainty, under the assumptions made, is the Extended  Bayes' method, where uncertainty is
completely described with an extended probability distribution.
With the additional Robustness assumption, uncertainty is
completely described with a set of such distributions.
\end{Claim}

There are many results of the type above in the 20th century literature
on probability and uncertainty. Nevertheless, there is no consensus among
researchers in uncertainty management that Claim~1 is even
approximately valid. The Claim is of obvious interest as
a foundational issue. But interest is not confined to the ivory tower:
Designers of future complex systems that need new types of human-agent
and agent-agent interactions struggle with this question, and fielded
solutions have already unraveled difficult compatibility problems, e.g., 
when advanced system components using different, sometimes incompatible,
unknown or irrational
ways to describe uncertainty are
put together to form systems of systems. If Claim~1 above can be 
accepted, this task
becomes easier, since questions can center around requirements on useful
probability models rather than comparing disparate schemes of
uncertainty management.

The auxiliary argumentation required to support our claims
is similar to inference principles in statistics, like the Sufficiency,
Likelihood and Conditionality principles\cite{Robwass}. 
Such principles have only a pragmatic
validity, meaning that their acceptability may be questioned in new
applications different from those conceived when they were first accepted.
We will state a number of assumptions made in derivations of this kind,
and they are indicated in bold face.
Those that will eventually be retained in this study are marked with an
asterisk ({\bf *}).

In section 2  we review the basic features of plausibility spaces and
state a number of common assumptions used to analyze them. In section 3
we review the two basic methods used to derive probability as canonical
uncertainty measure, coherence arguments and Cox consistency based
argument. In section 4 we discuss a number of fundamental principles
that are also required but which are often hastily glossed over:
these principles have the common feature that they say that inferences
possible should not depend in an arbitrary manner on how our problem is
embedded in a larger context.
This discussion leads to Definition~1. The consequences of Theorem~1
are discussed in section 5. In section 6 we discuss extended
probability. Theorems are proved in the appendix.
The scientific contribution claimed is the concept of embeddability of
the auxiliary functions in an ordered field, and its proofs. These are somewhat more
complex than we hoped, mainly because the function $G$ is a partial function
on the domain of plausibility values.

There are two papers that analyze a similar question and lead to 
different but related and compatible  conclusions,
namely Kraft et al\cite{Intuitive} and Hardy\cite{Scaled}.

\section{Symbols and Plausibilities.}

The methods of definition and induction are attributed, by Aristotle, to
Socrates\cite{Barnes}, and were developed further by Aristotle and
the Stoic philosophers.
Although present English translations of  Aristotle contain words like
symbol, utility and probability, it is clear that Aristotle was not very interested in mathematics
and does not give quantitative models of decision problems.
So his terms do not
reflect a detailed and precise understanding of modern Bayesian decision
theory. In particular, the modern concept of probability was certainly
not known at the time. Nevertheless, Aristotle is the first to give a
surviving qualitative version of the Bayesian decision making principle in 
his Nichomachean ethics\cite{Aristethic}:
`` Find out what you think is a good life, and consider the probabilities of
your possible actions to achieve this. Then follow the course of action
which with highest probabibility results in a good life.''

Two thousand years later, Thomas Bayes was the first to apply the 
newly invented
probability calculus to an inference problem.
In his posthumous essay he computes the posterior probability for
the probability of heads in tossing a coin, with respect to observed outcomes
 and under the assumption that tosses are independent and identically
distributed, and that
the prior probability of getting a head 
is uniformly distributed between 0
and 1. Somewhat later, Laplace gave a more satisfactory analysis of the 
Bayesian method.

 Our modern plausibility
models assume a set of events, or statements, or possible world sets $A$,
$B$, $C$, {\em etc.,} which in an application will correspond to
conditions existing in the world or in the minds
of humans and agents. This step is completely standard.
Many theories of plausibility go on to define conditional event plausibilities,
which in many cases are   written using the probability inspired
notation $A|B$ - the plausibility of $A$ given that we know $B$ to be
true\footnote{$A|B$ is thus the plausibility, not the conditional event. It is sometimes written $pl(A|B)$}. This is the model in which foundational studies are usually made,
and which we use here.

There are essentially two different methods that were used to analyse possible
plausibility spaces for consistency. The more common is based on a gambling
analogy: plausibilities are assumed analog to betting odds, 
and a situation where a gambler can pick bets on a set of related events to hedge completely the risk of loosing money is called incoherent. The plausibility assignment leading to such a situation
is considered incoherent, and under various assumptions several arguments exist that
end up in concluding that probability or a set of probability distributions is the only coherent plausibility space
\cite{Savage,Lindley,Finetti,Shafervovk}. We will use here the alternative approach 
devised by Cox\cite{Cox46} which avoids the betting scenario and instead analyses how plausibilities of combined events can be defined, and what properties the plausibility combination functions must have in order to avoid that
an events plausibility will depend on which of several possible derivations is used.

Cox\cite{Cox46} assumes that the domain of
plausibility values is an interval of real values,
which without loss of generality can be assumed to be $[0,1]$, with 0 for
falsity and 1 for truth. He furthermore finds that two functions $F$ and
$S$ must exist such that $A\land B|C=F(A|B\land C, B|C)$ and
$\lnot A|C=S(A|C)$. With a number of -- sometimes implicit -- 
regularity assumptions,
he shows that the domain of plausibility values can be rescaled so that
$F$ is tranformed to multiplication, and $S$ to the function $S$ such that
$S(x)=1-x$. In other words, if his assumptions are accepted, plausibility
models must be equivalent to probability models. We can state an
assumption
made implicitly by Cox here (the function $G$ was introduced by 
Acz\'el\cite{Aczel66} in a related investigation):

{\bf Sufficiency assumption*:} The plausibility value of a proposition is 
a sufficient characterization of the statement with respect to
uncertainty and propositional connectives, {\em i.~e.,} there are
functions $F$ and $S$, and a partial function $G$ such that $A\land B|C=F(A|BC,B|C)$,
$\overline{A}|C=S(A|C)$ and  $A\lor B|C=G(A|C,B-A|C)$.

There has been some concern with alternative interpretations of symbols
used. In rough and fuzzy set theory\cite{Pawlak,Zadeh:1997:RFL}, the 
meanings of symbols are taken
to be ambiguous, in the sense that even with full information we cannot
definitely say, {\em e.g.,} whether or not $A$ obtains or if $x$ is a member of
set $X$. The ambiguity  problem
has in many articles been claimed to be an obstacle to probabilistic
uncertainty management\cite{Walley}. 
An alternative way is to see ambiguity
as an aspect that should go into the structure of a probabilistic model.
So instead of introducing $A$ as an objective fuzzy set, we can introduce
the judgments made by various agents and decision makers as $A|C_i$
for the judgment made {\em e.g.,} by agent $i$ based on the information
and background factors available to agent $i$. This makes fuzziness and
plausibility orthogonal concepts, and both can be present in an
application. Those who focus on impreciseness assume that this is
the most prominent characteristic of their applications, while those
focusing on plausibility do not think so.

Which is the domain $D$ of plausibility values? Fine\cite[Ch 1]{fine} lists
a number of theories, which correspond to different order
structures of the domain of plausibility values. In short,
the domain can be discrete (with true/false as the case equivalent to
propositional logic), ordered, continuous or partially ordered.  All
applications of plausibility in real systems aim at decision making,
and for this reason we must eventually be able to say that one event
is more plausible than another one. 
 For
this reason it is essential to have some order structure in the domain,
and we rule out completely unordered domains like the complex numbers.
This means that the relation 'more plausible than'â is transitive. This is
not necessarily the case in practical decision making but because of psychological effects like framing, unstable preferences or computational limitations of the brain.
Our study aims at prescriptive theories so this problem will be ignored: the domain of plausibility values is always (at least) partially ordered and
usually linearly ordered.

There are several uncertainty management methods that make use of
partially
ordered domains of plausibility values, typically in the form of
intervals of real numbers. 

{\bf Robustness assumption*:} A plausibility space with a partial order
relation on the domain of plausibility values, describing the relation
'more plausible than', is definable using indexed sets of plausibility
spaces each having a linearly ordered domain of plausibility values.
In such a space, an event $e_1$ is more plausible than $e_2$ if and only if it is more plausible in each constituent indexed space.

We have not yet a compelling argument for the robustness 
assumption\footnote{see however the restrospective section 8.2} --
existing proposals have rather large gaps. A possible
approach is to develop a general theory of 
partially ordered plausibility spaces.
Until this has been done, we assume robustness and consequently analyze  models with a
linearly ordered domain $D$ of plausibility values.

{\bf Monotonicity assumption*: } The domain of plausibility 
is  ordered,  $S$ is 
decreasing\footnote{Decreasing is a stronger condition than non-increasing. The assumption could also have been called strict monotonicity.
Despite the subtle distinction, the weaker assumption would lead to completely
different conclusions\cite{identical} since non-strict monotonicity does not entail cancellation laws.}, and $F$ and $G$ are
increasing   in each argument (in the case of $F$ if the other argument is
non-$\bot$). 

By considering the limiting case of propositional logic we also find
a number of constraints based on the requirement that reasoning
with entirely false or true events should follow the rules of propositional
logic. We omit a number of rules which will follow from the algebraic rules of
auxiliary functions $F$ and $G$ that we will soon adopt:

{\bf Propositional limit assumption*:} $G(x,\bot)=x$,
$F(\bot,x)=\bot$,
 $F(x,\top)=x$, and $S(\bot)=\top$. 

An immediate consequence of Monotonicity and Propositional limit is $F(x,y)\leq \min(x,y)$ and $G(x,y) \geq \max(x,y)$.

\section{Coherence, or Dutch Book avoidance}

We have now stated the common assumptions of most existing theories of
plausibility\cite{fine,Finetti,Cox46,Savage}. But more is required before
probability appears as inevitable in a form we recognize. What is
required is some means to derive constraints on sets of plausibilities
of different conditional statements.
Without such constraints we can easily define plausibility spaces
that are not equivalent to probability. There are basically
two ways to proceed. One followed by de~Finetti is to construct a gambling
situation where different plausibilities are tied together in composite
bets offered by a bookmaker according to his beliefs. A bookmaker who
offers a set of bets among which a gambler can choose a combination
which gives positive payoff in every situation seems to have made
his bets from a globally incoherent set of beliefs. Several papers show,
with important differences in detail, that every coherent belief set is
equivalent to a
probability model\cite{Finetti,Cox46,Savage,Lindleya}. 
The coherence
concept can also be applied to some domains not totally but partially
ordered, and Goodman, Ngyuen and Rogers\cite{Goodman} show that some plausibilty measures not coherent under the
total order assumption, among others the standard version of DS-theory, are coherent under the latter assumption.

Another way to connect different plausibilities is via propositional
logic connectives, and this gives us some constraints on the
auxiliary functions $F$, $S$ and $G$ used in Cox's approach.
As an example, the rule of associativity for conjunction 
enforces $F$ to be associative for certain arguments,
namely, from
$
ABC|D=F(AB|CD,C|D)=F(A|BCD,BC|D) 
$
and $x=A|BCD$, $y=B|CD$ and $z=C|D$, we can derive

\begin{equation}
F(x,F(y,z))=F(F(x,y),z).\nonumber
\label{eq:assoc}
\end{equation}

 In other words, under some circumstances associativity is
inherited by $F$ from the associativity of conjunction.
If we want to show that every good plausibility space must be
rescalable to probability, we must prove that $F$ and $G$ satisfy certain
algebraic laws that are satisfied by $\cdot$ and $+$ and invariant under
rescaling - the laws of associativity, commutativity and distributivity
satisfied by the operators of a ring.
As an example, a model is discussed by Halpern\cite{Halpern} that has only a 
few 
events. Thus the domain of plausibility values is finite and associativity
and differentiability of $F$ does not seem inevitable. In order to
claim that the auxiliary function must be associative, we must either
claim that no sensible person would deny it, which is difficult,
or find some type of argument that goes outside the finite example.
In \cite{Cox46}, it is plainly assumed (as pointed out in \cite{Snow})
that the auxiliary function $F$
has certain regularity properties:

{\bf Associativity and differentiability assumption}:
The function $F$ is defined on $[0,1]^2$, and is associative and twice continously
differentiable; the auxiliary function $S$ is continuously
differentiable and $S(S(x))=x$

The differentiability part is implicit -- as was common when Cox wrote his paper -- but is a standard assumption made to
justify switching the order of differentiation which occurs in the derivations
of \cite{Cox46}, and
also in \cite{jaybook}.

\section{ Embeddability, Denseness, Refinability.}

The argumentation reviewed in section 3 must be complemented by some arguments
that bind together plausibilities of events that are not connected by 
propositional identities. In Cox's work, this is effected by regularity
assumptions: the auxiliary functions $F$ and $S$ are assumed to be defined on
an interval of real numbers, and obey the associativity and differentiability
assumption on this interval. Implicit in this assumption seems to be the
idea that the functions $F$ and $S$ are universal, and every plausibility
problem should be solvable with the same functions $F$ and $S$. In the
criticisms of Cox's work, this assumption has either been assumed not to 
exist\cite{Halpern, halpernta}
or been ignored\cite{paris}. The statistics based derivations of  probability
as universal uncertainty measure often mention similar assumptions,
but surprisingly often these are somewhat glossed over, the main exceptions
being \cite{BS,fine,Savage,Walley}. We make a sketchy review of some of the
assumptions that have been proposed. In the prevision-based analyses, the 
first methods were based on assuming that an arbitrarily fine partitioning 
of the events space exists; this allows one to deduce the existence of a
probability measure based on a relation $\leq$ on events  with the meaning
$A\leq B$ if $A$ is equally or less plausible than $B$.

{\bf Uniform partition assumption}\cite{Finetti}:
For every event $B$, and every number $n$, there is a partition of $B$
into $n$ equally plausible events.

{\bf Almost uniform partition assumption}\cite[Ch 3]{Savage}:
For every event $B$, and every number $n$, there is a partition of $B$
into $n$  events, such that for $r=1,\ldots, n-1$, the union of any
$r$ elements of the partition is less plausible than the union of
$r+1$ members. 

{\bf Standard events and precise measurements assumptions}\cite[Ch.~2.3,
Ax.~4,5]{BS}: For every conditional event there exists a standard event
with the same plausibility, and for every number $r\in [0,1]$ there is a
standard event with probability $r$.

{\bf Embeddability assumption}\cite{Walley}: Inference should not depend
on how the events participating in calculations
are embedded in a larger event plausibility
system. 

An alternative assumption used to 
derive Cox's result is given by Paris\cite{paris}:

{\bf Denseness assumption}\cite{paris}: For all real values $x$, $y$, $z$
and each
$\epsilon>0$, there are events $A$, $B$, $C$ such that $|(A|BCD) - x|$, 
$|(B|CD)
- y|$ and
$|(C|D) - z|$ are all less than  $\epsilon$.

This assumption might be more motivated than Cox's original
assumptions, although it does not seem to be weaker.
 In response to \cite{Halpern}, we developed very
weak assumptions binding the different parts of a system of events
together\cite{ASEC,ME2000}.

{\bf Refinability assumption*}\cite{ASEC}: In a plausibility model with
a conditional event of plausibility $p$, 
it must be possible to introduce a new subcase $B$
of a non-false event $A$  with  plausibility value $p$ given to
$B|A$. This means that it should also be possible to define a new subcase of
an event and get a new model which is equivalent to the original one as
long as no reference is made to the new event. If two new subcases $B$
and $B'$ of an event $A$  are defined in this way,
they can be specified to be information independent, {\em i.e.,}
$B|B'A=B|A$, $B'|BA=B'|A$. For two plausibility values $x$, $y$ such that
$x<S(y)$, it should be possible to define two new subcases $C$, $C'$ of any
non-false event 
$A$ such that $x=C|A$, $y=C'|A$ and $C\land C'|A = \bot$.

These augmentations of the model are called {\em refinements}.
A refinement just introduces an already existing plausibility value in
'another part' of the model. This means that there is no assumption of
a 'dense domain' or that a given new plausibility value can be introduced.
The motivation for refinability is that the same cognitive or other
process that
resulted in a particular plausibility value for one event can always
be mirrored in another part of the model to produce the same plausibility
value for another event. If this process introduces an inconsistency, then
we feel that there is a basic shortcoming of the model, and we should not
accept it. We claim that this assumption is weaker than those
referenced above in this section, and also more motivated.
This is the weakest condition we have found that ensures
that the auxiliary
functions $F$, $G$ and $S$ have enough algebraic properties
-- associativity, symmetry({\em i.e.,} commutativity for corresponding binary operators)  and distributivity properties
of + and $\cdot$ in a ring -- to ensure
that the main theorem follows.
Somewhat surprisingly, Cox's result on
rescalability of the plausibility space to standard probability follows
from refinability and the assumptions in section~2 for models that have a
finite number of plausibility values and an ordered plausibility 
space\cite{ASEC}. 
Even more surprisingly, the
same result cannot be proved for infinite plausibility spaces
(non-provability follows from a counter-example\cite{ME2000}). Two
responses to this are possible: either more assumptions are
required, or the result would not be right. We choose the second
alternative. The reason is that we want to extend probability to extended
probability, where infinitesimal probability values are allowed.
This idea goes back to Adams\cite{Adams} and has been found to give a 
very basic uncertainty management that
seems to incorporate many uncertainty calculi
as special cases (several are analyzed using extended  but not necessarily Bayesian
probability in \cite{DuPr}). The combination of robust and extended 
probability has been analyzed by Wilson\cite{Wilson}. 
Our motivation for accepting extended
probability is the existence of infinite plausibility models that are
refinable and embedded in the
real numbers, but whose plausibility spaces are  
not rescalable to standard probability\cite{ME2000}.

In order to derive extended probability as a canonical plausibility space,
we must
assume that models can be refined to the limit:

{\bf Closedness assumption*}\cite{ME2000}: The functions $F$, $S$ and $G$
have the following additional properties: 

$F:D\times D \rightarrow D$,\hfill\break
$S:D\rightarrow D$, \hfill\break
$E=\{(x,y)\in D\times D:x\leq S(y)\}$ and
$G:E\rightarrow D$.

In other words, the functions $F$ and $S$ are total while $G(x,y)$ 
is defined when $x\leq S(y)$.  This requirement is similar (but not equivalent) to Kolmogorov's
insistence that a probability space is defined on a $\sigma$-algebra.
Refinability would imply that $F$ and $G$ must be associative and
symmetric, and also that $F$ must distribute over $G$ on the domain of
plausibilities (since $G$ is a partial function, we must moderate its
laws by only requiring that both sides of its associativity and symmetry
equations are equal when one of them has a defined value). These algebraic
properties will simply be inherited from the corresponding properties of
$\land$ and
$\lor$. We can now see that Definition~1 of a proper
ordered plausibility space is relevant in the sense that  we must
work with such spaces if we accept the general framework of section 2,
and if we also accept the refinability and closedness assumptions. If we
accept the Robustness assumption we can concentrate on ordered spaces.

\begin{Definition}
A {\bf proper plausibility space} is a seven-tuple
$(D,F,G,S,\leq,\bot,\top)$, where $D$ is a partially ordered domain with
smallest value $\bot$ and largest value $\top$, $F:D\times D
\rightarrow D$, $S:D\rightarrow D$,
$E=\{(x,y)\in D\times D:x<S(y)\}$ and $G:E\rightarrow D$. Moreover,
$F$ and $G$ are symmetric and associative, $F$ distributes over
$G$, $F$ and $G$ are increasing in their arguments and $S$ is
decreasing. Additionally, $G(x,\bot)=x$, $F(\bot,x)=\bot$,
 $F(x,\top)=x$, $S(S(x))=x$, and $S(\bot)=\top$. 
A {\bf proper ordered plausibility space} is a proper plausibility
space where $\leq$ is a total ordering.
\end{Definition}
 
 \begin{Theorem}
A proper ordered plausibility space can be uniquely embedded in a
minimal ordered field where multiplication and addition are extensions of
$F$ and $G$, respectively.
\end{Theorem}
{\em Proof.} See appendix.

\begin{Theorem}
Assume that a plausibility model and its plausibility space satisfies
the assumptions of Sufficiency, Monotonicity, Propositional limit,
Refinability and Closedness. Then the plausibility space is a
proper plausibility space.

\end{Theorem}
{\em Proof.} See appendix.

We can now see that Theorem~2 leads us to concentrate on proper
plausibility spaces, and that with the Robustness assumption 
Claim~1 should follow from Theorem~1.

\section{Common sense assumptions entail extended robust probability.}

The best known ordered fields are ${\bf Q}$ and ${\bf R}$.
However, there are ordered fields that contain more than the real
numbers. One example is the field of rational functions $R(\epsilon)$ in
one variable $\epsilon$. These functions can be added to and multiplied with
each other, and if they are ranked by the lexicographical ordering of their
values and derivatives of order 1, 2, ... at $\epsilon=0$, we get an ordered field. This field contains all real
numbers (as constant functions) but does not admit lowest upper and
greatest lower bounds, and  thus cannot be embedded in the field of
reals. The variable $\epsilon$ can be regarded as an infinitesimal,
positive but smaller than every positive real number. There
are many ordered fields that are superfields of the reals, and there is
even a unique maximal ordered field {\bf No}, described by
Conway\cite{Conway}. This field consists of reals, infinite numbers and
infinitesimals. There is an infinitesimal element (a non-zero element
smaller than every real number) for every infinite ordinal and
transfinite number, namely its inverse. The field {\bf No} has
an extremely complex structure 
and cannot be easily represented in a computer. The concepts of
infinitesimals and infinite numbers in non-standard analysis\cite{Arob} are
closely related to extended probability. However, non-standard analysis has a
set-model-theoretic basis  whereas our basis is classical algebra,
and it is not completely
obvious that the two concepts are exactly identical.

The above leads us to a belief structure where belief is an extended
probability value. Many arguments have been
forwarded for belief to be modeled not by an ordered set, but a partially
ordered set like a set of intervals. Because of the subtle properties of
belief, it is difficult to resolve what the right thing is in a
convincing way. The preceding analysis started out with the assumption
that belief values are totally ordered, and cannot get to other conclusions.
However, assume that the possible plausibility values form a partially
ordered set, and accept the Robustness assumption of section~2. This
partially ordered set can then be regarded as a set of different contexts
({\em e.g.,} different unfused expert assessments) of linearly ordered
plausibility models. For each linearly ordered domain belonging to an indexed member , Theorem~1 is applicable,
and thus a plausibility model can be fully characterized by a set of
extended probability distributions, one for each index.
One way to fuse such sets of distributions is to form a weighted average of the distributions, but it is not clear which principles such fusion should be based on.

\section{Pragmatics}

If we design a methodology where users are permitted to define epistemic
states as sets of extended probability distributions we will soon find
us in a situation where these users are encouraged to define things
they cannot possibly understand. Extended Robust Bayes allows us to
construct extremely complex descriptions of systems,
where instead of points we must work with polytopes in high-dimensional
or even infinite-dimensional spaces. Even simple things cannot easily
be done with such models\cite{A1}. The most interesting aspect of the above
analysis is that it indicates a possible maximal generality in uncertainty
management, but in every particular application this generality must probably
be pruned.  Is the full generality really required? There are strong
claims in the literature that it is not. There are also strong claims that
conventional Bayesianism, where only one probability distribution is used
and where probabilities are standard, is inadequate. So, a pragmatic
approach would be to investigate claims made, and see how reasonable they 
are. In this investigation  there seems no definite need to distinguish
human from machine decision making -- most requirements we find
are obviously applicable to human and machine alike, particularly when we
want machines to interact with humans in terms of beliefs and intentions.

\subsection{Extended Probability}

Is extended probability needed or is standard probability adequate? 
Extended probability is not completely
unavoidable, in the sense that every
finite extended probability model is equivalent to a
standard probability model\cite{ASEC}. But in some cases extended probability
seems useful as a pragmatic simplification of a modeling problem and obtaining
natural problem descriptions. It seems clear
that there is a phase in cognitive assessments where qualitative and
order-of-magnitude reasoning is done\cite{natdes}, followed by a quantitative
phase where more quantitative reasoning occurs.
In AI reasoning research, many non-probabilistic methods have
been proposed, and it seems as if many of them
can be described in terms of extended probability\cite{DuPr,Sme}.
Non-standard analysis approaches have been advocated as an
alternative to measure-theoretic ones in stochastic processes.
Although no immediate reaction followed from the user communities,
this is still a promising direction\cite{Nelson}. A number of assumptions
have been proposed in the literature for arriving at the
inevitability of standard probability and thus the exclusion of extended
probability:

{\bf Real valued assumption:} Assume (with, {\em e.g}, Cox)  that plausibility is real valued. 

{\bf Archimedean assumption:} Fine\cite{fine} assumes that for every non-zero
probability $e$, with $ne=G((n-1)e,e)$ and $1\cdot e=e$, there is an $N$
such that $Ne>S(e)$. This assumption is introduced to step from comparative
to standard probability.

{\bf Separability assumption:} Arnborg and Sj\"odin\cite{ME2000} define
a separable model as one in which, for every $x<y$ and $c$,
there are $n$, $m$ such that $x^n<c^m<y^n$, with $x^n=F(x,x^{n-1})$ and
$x^1=x$. This is introduced as a weaker assumption than continuity of the
auxiliary function $F$ introduced in \cite{Aczel66}.
 
With the hindsight given by Theorem~1, these assumptions appear only as
alternative  ways to say that we do not accept infinitesimal probabilities.
None of these assumptions are very compelling except possibly the first,
if plausibility values are used in an application strictly using expected utility 
decision making. In this
case plausibilities separated by an infinitesimal amount cannot be
distinguished and can be considered equivalent.

\section{Summary and Conclusions}

Many of the objections to Bayesian methods come from
the sometimes very complex analytical models preferred by
theoretical statisticians, and many practitioners have seen alternative
methods like neural networks as a way to by-pass statistics. 
Unfortunately (or fortunately) this is to a large extent an illusion.
There is no principled reason why models used could not be based
on other types of models like neural networks, linguistic coding or
case based reasoning, and indeed efforts have been made to view these
techniques as special types of Bayesian 
models\cite{IJCAI99-Vol1*248,RipleyNN}. 
As detailed above, there is no simple way around the normative claims of 
the Robust Extended Bayes' method. So if one uses an alternative method,
it will sooner or later have to be evaluated against the standard
scale of rationality. Bayesian methods contain a large amount
of freedom in the sense that there are no 'correct' models or
model sets on which to base conclusions about real world or inner world
phenomena, but models must be chosen and tested against requirements of
applications. This makes the Robust Extended Bayesian method itself 
more or less impossible to
falsify - but so are other very basic methods like arithmetic. 

Is it now possible to answer the question in the title of this paper?
Even a subjectivistic probabilist would hesitate, but mainly because the
question is a little ambiguous. Our answer is
that extended robust probability is completely plausible as a 
universal uncertainty or
belief measure until an example is given where some of the starred
assumptions above can be demonstrated dubious.

\section*{Acknowledgments}

The reviewers of previous versions of this contribution have 
influenced it significantly, by 
pointing out questionable statements (some now deleted,
others elaborated) and
several relevant related papers in many disciplines.

SA is indebted to David Draper for discussions of the importance of and approach to including countable additivity in the Cox/Jaynes framework,
which also made me finalize this unpublished manuscript and add a retrospective section below (after the bibliography).

\bibliographystyle{plain}
\bibliography{mine}

\section{Retrospect}

Various versions of this manuscript were accidentally indexed by search engines and 
are cited by other papers. This is the last 2003 version, with very few changes made in 2015: typos removed and a few
too short arguments elaborated.  Nothing reflecting developments since 2003 has been 
added, except in this retrospective section. Several aspects of possible polishing of Cox/Jaynes's
development have been published, for a survey see the introductory sections of \cite{TereninDraper}.
Apparently, little effort has been spent on analysing extended probability and partially ordered plausibility domains, but there are a few results to review. 
I will only mention those results easily discussed in our algebraic framework. Some papers argue that we have made above quite a number of assumptions
compared to other papers in the same area. But this is, on closer reading, only an effect of our need to be explicit about assumptions and to discuss several
alternative assumption sets. Counting assumptions as axioms is a bit misleading since we actually want compelling axioms, not minimum number of them.
The shortest axiom set would be  our Claim above... 

\subsection{Countable additivity}

In the paper by Terenin and Draper\cite{TereninDraper},
a new aspect is taken up, namely countable additivity of probabilities which has several equivalent definitions, the one
coupled to its name is that for all events C and all sets of mutually disjoint events $\{E_i\}$, we must have $\Sigma_i P(E_i | C)=P(\bigcap_i E_i | C)$.
Yosida and Hewitt, however, define countable additivity as the condition that for every event $C$ and sequence $(A_i)$  such that $A_{i+1}\subset A_i$ and
$\bigcap_i A_i =\emptyset$, we also have $\lim_i P(A_i|C) =0$, and they claim it trivially equivalent to the former definition. The latter seems more
understandable than the first, and could possibly serve as a 'compelling assumption' in the Cox/Jaynes framework, as is approximately the case in \cite{TereninDraper}.

Although one normally, in applications where the question arises, just assumes or postulates that this assumption is fulfilled (as among others Kolmogorov did), it actually does not
have to be true for infinite sets of events (but it must be true when  $\{E_i\}$ is a finite set, a property called finite additivity), this being a consequence of somewhat counter-intuitive measure-theoretic considerations. A simple example of a probability distribution violating countable additivity is
obtained by starting with a uniform distribution over integers  $1$ to $N$ and letting $N$ go to $\infty$. Then every probability of an integer goes to $0$ but
all the sums are $1$. The limit of the sums is thus $1$ while the sum of the limits is $0$. One common way to handle this is to claim that the limit, an improper prior, is not a probability distribution, despite the fact that unnormalized improper priors are sometimes (actually, quite often) used 'as if' they can be normalized. If the likelihood is enough concentrated, the posterior obtained by applying Bayes rule can be normalized and thus gives an inference, although it 
is clear that there are some problems with this approach. 
An improper prior, however, contains more significant information than the normalized finitely additive probability distribution, as can be seen in that the standard improper priors (uniform, Jeffrey's, etc) are identical as probability distributions: they all have density zero (except, for Jeffrey's prior, at zero). However, in the framework of extended probability they are proper distributions with  infinitesimal densities. The question whether probabilities
in infinite spaces should be countably or just finitely additive has been debated a lot, while in most applications one just assumes, with Kolmogorov, countable additivity. The Cox/Jaynes framework is explicitly based on what Jaynes refers to as common sense, but the question of countable/finite additivity has not previously (before Terenin and Draper) been analysed within this framework. Both Cox and Jaynes most likely regarded this as a non-issue, and for non-parametric Bayesian inference it is clear that current practice is based on the assumption of countably additive probability. Considering the current situation, we have here a case were researchers have not quite agreed about what common sense requires. From an engineering perspective the most difficult to swallow property of spaces not fulfilling countable additivity
is known as non-conglomerability (de Finetti\cite{Finetti}). Non-conglomerability can be defined in terms of plausibilities: For an infinite partition  $\{E_i\}$  of $E$ and an event $C$, it is not necessarily the case
that $\inf_i C | E_i \leq C | E \leq \sup_i C | E_i$. Although my own reaction to this, both immediately and after considerable reflecting over it, 
is that non-conglomerability is absurd and must be rejected, the analyses of consequences of non-conglomerability are sometimes just mentioning
that the concept invalidates some familiar inference methods. For example, it is not permitted to conclude that $C|E$ is a weighted average 
of the $C|E_i$ for an infinite partition $\{E_i\}$ of $E$, in other words among the $C|E_i$ we will not necessarily find both numbers not less than and numbers not greater than $C|E$. Is my reaction to this a sound reaction or just unwillingness to change habits?

If  non-conglomerability is rejected and conglomerability is accepted as a common sense assumption, we can only end up with a system of plausibilities that can be rescaled to a system of probabilities that has to be conglomerable. Although it is relatively easy to see that a countably additive probability space is conglomerable, the reverse is also true but not so easy to see: a conglomerable probability space is countably additive. This was proved in \cite{Schervish},
so conglomerability entails countable additivity, at least for real valued probability systems.  This is an alternative axiom to that presented in \cite{TereninDraper}, maybe also more compelling.
What happens in extended probability has to be investigated, 
since the existing derivation relies in many places on standard probability, where, for example, least upper bounds exist. 

\subsection{Partially ordered domains and robustness}

Concerning partially ordered plausibility domains, little progress has been made. However, J\"org Zimmermann in his thesis \cite{ZimmerTh}
 shows that a partially ordered plausibility space is embeddable in a partially ordered ring, and shows, assuming that this ring has a greatest ordered subfield called backbone (and satisfies additional technical constraints), that any plausibility value $r$ can be uniquely decomposed into $s+a\cdot t$, where $s$ and $t$ are from the backbone and $a$ is interactive, which means that $a$ is incomparable to elements between 0 and 1. He also shows that DS-theory is interpretable in this way and clearly robust probability also fits the model.
In robust probability the plausibility domain is a vector (or indexed set) of probabilities and the backbone elements have all components equal. The domain is a ring, not a field, since it has zero divisors. The interpretation of the decomposition here is that $s$ is an estimator of the probability (actually the min probability), $t$ is a measure of the 'uncertainty of the probability' (maybe $s+t/2$ is a better estimator of the probability), and $a$ gives a kind of 'profile' characterizing the deviation of the uncertainty from the estimated probability. This is quite nice, but the assignment of the estimate to every plausibility in the model does not itself give a coherent probability assignment. It is also not clear that the mentioned subfield always exists and thus a general understanding of partially ordered plausibility
has not yet been obtained, in particular we do not know what weight the robustness assumption carries. Zimmermann, in his ring theorem, clearly makes an assumption slightly stronger than we have done, in the axiom $\mathbf{And}_2$. This entails the cancellability law for + of a ring, and we did (probably) not make enough assumptions to get cancellability. His assumption is clever and plausible, how compelling it is I cannot yet quite see.  We could prove cancellability for $G$  only for totally ordered plausibility domains.  Without some type of cancellability assumption it is difficult to characterize the possible plausibility domains, and in particular to develop the algebraic approach. Even with the assumption it is difficult to get a real grip on what the most general such domains mean in practical terms. Ring structures are quite flexible.
There is a fair number of in-depth treatments of partially ordered rings, the more recent motivated by its relevance to real algebraic geometry, but as far as I know they have not yet been seriously applied to the philosophical study of plausibility. It seems as if Zimmermann's  assumption ($\mathbf{And}_2$)
is a good starting point, and by his ring theorem the plausibility domain is then a partially ordered ring.  He calls rings suitable for plausibility calculus $c$-rings. Now, the $c$-ring does not have to contain a field
(the backbone) under assumptions made, since the ring of integers is a counterexample, and powers  (repeated products) of this ring are others. 
But if  there is a (non-integer) rational number (a rational number is a ring element $f$ such that the $n$-fold sum of $f$:s equals $m$ for some integers $n$ and $m$) in the $c$-ring it must also contain
lots of rational numbers since the ring is closed under addition, subtraction and multiplication and this subring of rationals seems a good approximation to a field, the backbone. Moreover,  any ordered field can be added to any partially ordered ring, creating a $c$-ring with a backbone from a $c$-ring without one.
So the decomposition theorem of Zimmermann holds for all $c$-rings, even if some $c$-rings may not have interesting decompositions.
A second question raised by the referenced thesis is the structure of $c$-rings. Our robustness assumption can also be formulated as a conjecture:

{\bf Conjecture:}
Every $c$-ring is embeddable in a product of ordered fields. 

Considering the intense studies made of rings and fields, one would expect the literature to contain the verdict on this conjecture. And it seems to do,
but the standard algebraic operating procedures makes it more appropriate to say that every c-ring is a subdirect sum of fields:

\subsubsection{Representing rings as subdirect sums}

The problem of representing a complicated ring as a product of simpler rings has been thoroughly studied in \cite{Birkring} and  \cite{Mc38,Mc47,Mc64}. A {\em product ring} $P=S\times T$ of two rings $S$ and $T$ has a carrier that is the Cartesian product of those of the factors $S$ and $T$, and the operations are defined component-wise, thus, for example, if $S$ contains $s_1$ and $s_2$ and $T$ contains $t_1$ and $t_2$, then
$P$ contains $(s_i, t_j)$ for $i=1,2$ and $j=1,2$. Moreover, $(s_1,t_1)\cdot (s_2, t_2)=(s_1\cdot t_1, s_2 \cdot t_2)$ and 
$(s_1,t_1) +  (s_2, t_2)=(s_1 + t_1, s_2+t_2)$. Two observations: for some, probably historical, reason, our main tool is actually a product
 but is called subdirect sum. Another important thing to note is that the product construction always introduces zero divisors, so the product of two integral domains is never itself an integral domain: $ (s, 0) \cdot (0, t) = (0, 0)$ regardless of $s$ and $t$, and $(0, 0)$ is the zero element of the product $P$ since it is the unit for 
 $+$. Therefore, a cancellation law for $\cdot$ never holds in a product or subdirect sum. By a {\em subdirect sum} S of rings
$\{S_i\}_{i\in I}$ (where the index set $I$ is not necessarily finite or countable) we mean a ring isomorphic to a subring $P$ of $\prod_{i \in I} S_i$, and such that
the projection of $P$ on component $i$ is exactly $S_i$ (i.e., the homomorphism is onto). Moreover, we require the representation to be {\em non-trivial},
i.e., none of the subrings $S_i$ is allowed to be isomorphic to $P$. 
If such a non-trivial representation exists, the ring is called {\em reducible}. If no such representation exists, the ring is {\em irreducible}.
Note that an irreducible ring is not the same as a reduced ring; the former can not be split into a subdirect sum whereas a {\em reduced} ring 
is a ring that has no nilpotent elements.

{\bf Theorem:} \cite{Birkring}
Every reducible ring is a subdirect sum of irreducible rings.

What makes this relatively simple theorem interesting is the connection to fields :

{\bf Theorem}\cite[Theorem 3.14]{Mc64}
Every subdirectly irreducible ring with no nilpotent elements is a field.

This is as close as we can get using the literature on rings to deriving the robustness assumption (In \cite[Theorem 4]{Mc38}, the corresponding statement was that every reduced (in the sense of absence of nilpotent elements) and reducible ring is a subdirect sum of integral domains, a slightly weaker result).  The question now is: is there some good argument to reject factors in
the subdirect sum  with nilpotent elements, which would recover the robustness conjecture? 
Yes, there is. If a factor in a subdirect sum $P$ has a nilpotent element then $P$ also has a nilpotent element and there is a nilpotent element in the plausibility domain:
If $S_i$ has the nilpotent element $a$, then the product has the nilpotent element 
$(0,\ldots,0,a,0,\ldots,0)$ with only one non-zero component. 
If there is a nilpotent plausibility value then there is also $a\neq 0$ such that $a^2=0$. Use the refinement assumption to define
two new subevents  $B_1$ and $B_2$ of $A$ such that $B_1|A=B_2|A=a$ and such that they are information independent: $B_2|B_1 A=a$.
Now $B_1$ and $B_2$ are both possible ($a\neq 0$) in the context of $A$, but their combination in the context $A$ has plausibility $a^2=0$ and is impossible. This 
contradicts the assumption that the events are information independent: since $B_2$ is possible in the context $A$ and information independent of $B_1$,
it should also be possible in the context $A  B_1$. We conclude that the ring is reduced, i.e., lacks nilpotent elements.
Since the decomposition of a reduced ring will result in a subdirect sum of fields, and these will be partially ordered by assumption,
we can use the fact(\cite{Schw04}) that a partially ordered field can always be totally ordered and thus is isomorphic to one of the fields between $Q$ and 
{\bf No}, to recover the robustness conjecture.

We have just informally proved:

{\bf Robustness Theorem}:
A plausibility calculus embeddable in a partially ordered ring is, under the refinability assumption, 
equivalent to a subring of a power $\mathbf{No} ^X$, for some set $X$. The domain of
this calculus is the set of maps from $X$ to subfields of $\mathbf{No}$ and the operations
$\cdot$ and $+$ are defined component-wise using the corresponding field operations.
 
The question remains of course how this set of probability assignments should be interpreted. Since the domain is a direct sum of ordered fields,
there is only one partial order to choose, where $a\geq b$ if and only if $a_i\geq b_i$ for all $i \in X$.

 The assumption is that event $A$ is more plausible than $B$ if each distribution in the index set says that $A$ is more plausible than $B$. Our assumptions do not let us conclude that the lower and upper envelopes have the specific meaning of lower and upper limits of an unknown probability distribution, although this interpretation is done in robust Bayesian inference. However, the domain values having all components equal represent obviously precise probabilities (as well as the backbone of \cite{ZimmerTh}) and
an event $B$ with imprecise probability (not all components equal) is more plausible than an event $A$ with precise plausibility if every component
of $B$ is more plausible than $A$, and thus an upper and by symmetry lower plausibility is indicated for the maximum and minimum components plausibilities.
The robust Bayesian interpretation of domain values is thus justified, and Zimmermann's decomposition $p=s+a\cdot t$ contains indeed the lower
and upper probabilities of $p$, namely $s$ and $s+t$.  There is a small complication in that sets of extended probabilities are sets of hyperreal numbers 
which do not necessarily have tight lower and upper bounds. The standard way to define infemum and supremum of such sets is to use the standard part of the numbers, although we have not investigated the appropriateness of this method in this particular application.

\subsubsection{Robustness examples}

Since the robust Bayesian representation seems universal one can ask why there are other uncertainty management schemes than robust Bayesian analysis,
or rather how these fare with respect to the Robustness theorem. The robustness theorem says that a problem where uncertainty is present should be 
modeled with a set of conditional probability assignments, and that these have the interpretation of possible probability distributions. This seems in agreement 
with how robust Bayesian analysis is performed and interpreted. Zimmerman considers also other uncertainty management schemes, and observes that the system of lower probabilities violates his ring theorem. This is because lower and upper probabilities are insufficient for recovering a set of distributions:
in the lower and upper envelopes of a set of distributions the latter are mixed up and there is no way to recover the set of distributions from which 
the upper and lower probabilities were generated. This has the consequence that the calculus cannot be embedded in a ring and there is no useful definition of conditional plausibility. This appears as a weakness or at least inconvenience of the method.

A third method, probably the most serious competitor besides Bayes, is Dempster-Shafer evidence theory, DS-theory\cite{Dempster,DS}.
Here, an assessment is described as a 'body of evidence' or mass assignment, and this is a random set over the frame of discernment $\Omega$ which
in turn is a partition of the universe of discourse. A generative model of the outcome given a body of evidence is that an outcome of the random set is generated and
then one member of the resulting set is chosen arbitrarily. If the random set has non-zero probability only for singleton sets there is no
arbitrary choice and we have a precise Bayesian probability model.
Zimmerman observes that a body of evidence can be translated to a set of distributions and this is a good beginning. 
However, the DS-theory also has the peculiarity that bodies of evidence that are independent (of which apparently no precise characterization has been 
given) should be combined with random set intersection (conditioned on being nonempty), the Dempster's rule. Dempster's rule is different from the 
robust Bayesian combination rule. A very useful example of the difference was presented by Gelman\cite{Gelman}: it involves a match between a boxer and a wrestler,
the outcome of which is entirely uncertain, and a coin flip which is a standard random event with probability 0.5 for each outcome. Here the events are:
$B$: boxer wins; $\overline{B}$: wrestler wins; $C$: heads up; $\overline{C}$: tails up. The frame of discernment is 
$\Omega=\{BC, B\overline{C}, \overline{B}C, \overline{B}\overline{C} \}$.  The total uncertainty of $B$ is expressed with the vacuous mass assignment
$[m_1(\{  BC, B\overline{C}, \overline{B}C, \overline{B}\overline{C} \})=1]$, a random set whose outcome is  always $\Omega$. The information on the coin flip is expressed by the body of evidence $[m_2(\{BC, \overline{B}C\})=0.5; m_2\{B\overline{C}, \overline{B}\overline{C}\})=0.5)]$, a random set whose outcome is $\{BC, \overline{B}C\}$ or $\{B\overline{C}, \overline{B}\overline{C}\}$, each with probability 0.5. Gelman now goes on and adds the evidence that an impartial and trustworthy person reports that the two events had identical outcome (either $BC$ or $\overline{B}\overline{C}$ happened) This can be encoded as the body of evidence $[m_3(\{BC,\overline{B}\overline{C}\})=1]$. Combining, using Depster's rule, the bodies of evidence $m_1$, $m_2$ and $m_3$, gives the
random set  $[m(\{BC\})=0.5; m(\{\overline{B}\overline{C}\})=0.5)$, in other words the outcome is that of the standard coin flip. On the other hand, if the 
bodies of evidence are translated to sets of distributions and combined using Laplace's combination (component-wise multiplication followed by normalization),
the outcome is completely uncertain between $BC$ and $\overline{B}\overline{C}$, in other words $[m_r(\{BC,\overline{B}\overline{C}\})=1]$, the random set 
outcome is always $\{BC,\overline{B}\overline{C}\}$ . 
There is a significant practical difference between these two results: In the first case any bet on the outcome at better than even odds is known to be advantageous, in the second case no bets or odds are known to be advantageous. And it is not difficult to see, by considering probabilities for $B$ close to 0 and 1, that in this simple example the robust Bayesian method  is the appropriate one. Gelman is dissatisfied with both results, but the robust Bayesian method actually gives the right result in this simple example, the difficulty seems to be in the problem formulation. Dempster's rule gives no imprecision when an imprecise body of evidence is combined with
a precise one (which, as mentioned, is characterized by being a singleton random set). This is a trivial consequence of the random set intersection
in Dempster's rule: the intersection of a singleton random set with any random set is itself a singleton or empty.

It is possible to bring in DS-theory under the robust Bayesian methodology, but only by modelling the random set operations directly -- they are after all entirely 
kosher probability models. This would entail introduction of new symbols for the subsets of $\Omega$. These symbols are new and do not correspond to unions or intersections but to the random sets involved, so for example the symbol $E=\{BC\}$ gives rise to the assignments $E|m_1=0$, $E|m_2=0$,
$E|m_3=0$ and $E|m=0.5$, whereas the correct answer based on $[m_r(\{BC,\overline{B}\overline{C}\})=1]$ would have $E|m_r=0$ but 
$F|m_r=1$ where $F=\{BC,\overline{B}\overline{C}\}$.
 This gives a precise Bayesian model where only the last step, translating the final random set to a convex set of probabilities,
uses the robust Bayesian framework. The problem is thus not that DS-theory does not fit into the Cox/Jaynes framework, but that the implied probability model sometimes 
(like in Gelman's case) gives wrong answers for simple problems, as has been pointed out in \cite{jaif,BrodzikEnders}. The discussion of the justification of Dempster's rule in \cite{Dempster} seems fair, but the problem is that the implied probability model is only known to be appropriate for problems where imprecision can
be attributed to private frames of reference.

\subsection{Philosophy}

A number of philosophical papers on the justifications of the line of investigations based on Cox's and de Finetti's work have also appeared. An article by John D Norton\cite{Norton} groups assumptions in several groups, gives counterarguments to some of them, and suggests that each group is checked and
a subset is chosen according to the needs of each specific application. This seems a reasonable idea, but it requires a lot of effort to find out what these subsets actually mean in terms of the ensuing uncertainty calculi, and my experience is that finding the specific needs in an application area is difficult, since as Norton points out the game is really to find convincing arguments for an already made decision. In practice it seems that, currently, a number of methods are developed in separate communities and one of their major tasks (given that the claim to universality is too difficult as they usually know by now) is to find the applications that confirm the method. The conclusion of Norton is quite plausible while his specific examples in the argumentation are sometimes difficult to follow and agree with, like the lack of distinction between descriptive (Cialdini,  Ellsberg, Kahneman, Tversky, Klein) and prescriptive studies. As an example, the non-associativity of the 'more plausible than' relation occurs only in descriptive studies (reflecting framing, cognitive barriers, unstable preferences or approximate computations) and there is no real argument against it in a prescriptive system. When and if a difference is found it seems that three 
lines of investigation are possible: (i) explaining the difference, for example by the wiring of the brain, the social context of the decision making, computational 
or cognitive limitations, or a real problem with the prescription; (ii) explaining what is wrong in practical decision making and trying to fix it by adjusting the education system; (iii) finding what is wrong with the prescriptive 
theories and adapting them to the real needs or finding a new prescriptive theory of some generality if not universality. It seems to me that accepting (iii) as the only appropriate way is not so convincing. The development of Bayesian analysis is however an example of the third option: from being harshly criticised most of the time before 1980, it has been tuned by an enormous effort by Bayesians addressing both fundamental and computational questions as well as developing sophisticated model families suitable for areas such as health sciences, bioinformatics, language translation, language and species evolution, robotics and vision, web analytics, signals and systems and finance, in several of which it has supplanted earlier methodologies. There is by now a toolbox that, by its diversity and the versatility of supporting computer codes, makes the Bayesian framework rather hard to replace.  It is also not the case that spectacular developments have been achieved all over the field and certainly some Bayesian efforts have faded away, but enough remains to make a convincing case for robust Bayesian analysis as the starting effort in an area, and in case the application area is resisting, see what modifications of  the method can be made before looking for completely different alternatives. Such modifications have been made, and I find it difficult to see many of them as contrived or exotic.

In summary, it seems that despite progress the last decade, there is more to find out about plausibility calculi.

\appendix

\section{Proof of  Theorem~1}

Recall the definition of an ordered field: it is a structure
$R=(D,<,\cdot,+,1,0)$, where $D$ is a domain ordered by $<$,
$\cdot$ and $+$ are total functions from $D^2$ to $D$, satisfying the
properties of associativity and symmetry, and where $\cdot$ distributes
over $+$. Moreover, it has an additive inverse, {\em i.e.,}, the
equation $a+x=b$ can be solved for $x$, and the equation $a\cdot x=b$
can be solved for $x$ if $a\neq 0$. The element 1 of $D$ is a unit for
$\cdot$ and  0 is a unit for $+$. The function $+$ is
increasing and $\cdot$ is increasing for arguments larger (by $<$) than 0.
Fields are special cases of rings and integral domains. An ordered
integral domain satisfies the rules of a field except that we do not
require solvability for $x$ of $a\cdot x=b$.

We will show how an ordered plausibility space is embedded in an
ordered field. 
We have
thus the  space defined by the ordered domain
$D$ with smallest and largest elements now called 0 and 1, respectively,
and they will always be embedded as 0 and 1 of the field, respectively.
Because of the significant number of algebraic
formulas used here, we also use infix notation $+$ and $\cdot$
for $G$ and $F$, as well as for their extensions.

The function $S: D\rightarrow D$ maps its argument $x$ to a solution 
- the only one because of the monotonicity assumption -
$y$ of the equation
$x+y=1$. The function $\cdot$
is defined on $D\times D$, and $+$ on 
$\{(x,y): (x,y) \in D\times D \land x\leq S(y)\}$. We now 
extend $D$ while extending the definitions of $\cdot$ and $+$.
We know that  $\cdot$ is associative and symmetric, and that $+$ 
is symmetric. If $(a+b)+c$ and $a+(b+c)$ are both defined, they are equal.
Likewise, if $a+b$ is defined, $c\cdot (a+b)=c\cdot a+ c\cdot b$,
and if $a+b$ is defined and $c<b$, then $a+c$ is also defined.
 
The strict monotonicity assumption leads to cancellation properties:
\begin{Lemma}\label{l:0}
If $a+b=a+c$ then $b=c$, if  $a\cdot b=a \cdot c$ and $a\neq 0$ then
$b=c$, if $a+b\leq a+c$ then $b\leq c$, and if  $a\cdot b\leq a \cdot c$
 and $a\neq 0$ then
$b\leq c$.
\end{Lemma}

If there are no elements in $D$ except 0 and 1, the embedding  
is trivial, so we 
assume an element $e$ with $0<e<1$ in $D$.
We do the embedding in three steps, using the standard technique of defining an
extension as the quotient of a set of pairs by an equivalence relation,
and indicating  which element of the extension that corresponds to each
element of the original domain. We use the notation $[a]_\sim$
for the equivalence class of $\sim$ containing $a$.
It is
easy, in each embedding step, to define the functions $\cdot$ and $+$ on
the extensions and verify that they are indeed functions and extensions,
that their laws are preserved, as well as to verify that no two elements
of the old domain become equivalent in the new domain. Details of this
verification are shown here, in the form of a series of Lemmas, 
whose proofs are sometimes terse and sometimes omitted.

\begin{Lemma}\label{l:3}
If $a_1+a_2$ is defined and $a_1 \geq b_1$ and  $a_2 \geq b_2$,
then $b_1+b_2$ is defined.
\end{Lemma}
{\em Proof:}
Obviously, $b_1 \leq a_1 \leq S(a_2) \leq S(b_2)$.

\begin{Lemma}\label{l:9}
If $e\neq 0,1$ and $f=\min(e,S(e))$, then $f\cdot a+f\cdot b$ is
defined.
\end{Lemma}
{\em Proof:}
Assume (no loss of generality because of symmetry of $+$) that $a\leq b$.
\begin{enumerate}
\item{$f=e\leq S(e)$:}
In this case, $f\cdot a\leq f \cdot b \leq S(f)\cdot b$, and
$f\cdot b +S(f)\cdot b$ is defined, thus by Lemma~\ref{l:0}
 $f\cdot a+f\cdot b$ is also defined. 
\item{$e>S(e)=f$:} 
Since $e\cdot b+S(e) \cdot b$ is defined and
$e\cdot b\geq f\cdot a$, thus by Lemma~\ref{l:3}
 $f\cdot a+f\cdot b$ is also defined. 
\end{enumerate}

\begin{Lemma}
For every sequence $(a_i)_1^n$
there is a non-zero $c_n$ depending only on $n$ such that 
$c_n\cdot a_1+c_n\cdot a_2 \ldots
c_n\cdot a_n$ is defined. 
\end{Lemma}\label{l:10}
{\em Proof:}
For any non-trivial plausibility
value $e$, choose $c=\min(e,S(e))$ and $c_n=c^{\lceil\log n\rceil}$.
Use Lemma~\ref{l:9} inductively on half sequences.

  The first embedding step
introduces non-negative rationals and is similar to the standard quotient
construction for integral domains. Let $D^+=D-\{\bot\}$ and
$D^{(1)}=(D\times D^+)/\sim$, where, for
$a,c\in D$ and $b,d\in D^+$,
 $(a,b)\sim(c,d)$ iff $a\cdot d=b \cdot c$. This makes $\sim$ an 
equivalence relation.
Use notation
$[a,b]$ for $[(a,b)]_\sim$. An element $d\in D $ is identified with
$[d,1]\in D^{(1)}$.

Define
$<$, $\cdot$ and $+$ as total functions on $D^{(1)}$ by
$[a,b]<[c,d]$ iff $a\cdot d<c\cdot b$,
$[a,b]\cdot[c,d]=[a\cdot c,b\cdot d]$, and
$[a,b]+[c,d]=[e\cdot a\cdot d + e\cdot c\cdot b ,e \cdot b\cdot d]$, 
for some $e$ 
such that the expressions are defined (see Lemma~\ref{l:10}).
The rational number $2/3$ is identified with
$[x+x, x+x+x]$ for some $0<x<c_3$, and the other non-negative
rationals are similarly defined. In this extension the rules for a field
are satisfied, except that we have not yet an additive inverse or
negative values: 
\begin{Lemma}\label{l:emb1}
The relation $\sim$ is an equivalence relation and
if $[d,1]=[d',1]$, then $d=d'$.
\end{Lemma}
{\em Proof:}
The relation $\sim$ is obviously reflexive and symmetric. It is also 
associative,
since if $(a,b)\sim (c,d)$ and $(c,d)\sim (e,f)$ with $b, d, f\neq 0$ then 
$a \cdot d=b \cdot c$ and $c \cdot f=d \cdot e$, hence
$c\neq 0$, $a\cdot f \cdot (c\cdot d)=e\cdot b \cdot (c\cdot d)$ and by the 
cancellation Lemma,
 $(a,b)\sim (e,f)$.

\begin{Lemma}\label{l:11}
The relation $<$ is  a total order on $D^{(1)}$.
\end{Lemma}
{\em Proof:}
Omitted.

\begin{Lemma}\label{l:12}
The relations $\cdot$ and $+$ are 
total functions on $D^{(1)}\times D^{(1)}$.
\end{Lemma}
{\em Proof:}
Consider $+$:
If $[a,b]+[c,d]=[f,g]$ and $[a',b']+[c,d]=[f',g']$ with $[f,g]=[f',g']$, then  
$[f,g]=[e\cdot a\cdot d + e\cdot c\cdot b ,e \cdot b\cdot d]$ and
$[f',g']=[e\cdot a'\cdot d + e\cdot c\cdot b' ,e \cdot b'\cdot d]$,
The equivalence condition  $[f,g]=[f',g']$ translates to
$(e\cdot a\cdot d + e\cdot c\cdot b)\cdot e \cdot b'\cdot d=
(e\cdot a'\cdot d + e\cdot c\cdot b')\cdot e \cdot b\cdot d$, in other words
(by cancellation) $(a\cdot d + c\cdot b) \cdot b'=
( a'\cdot d +  c\cdot b') \cdot b$
 so $[a,b]=[a',b']$.
 In other words the type of $+$ is 
indeed $D^{(1)}\times D^{(1)} \rightarrow D^{(1)}$.
The other cases are similar but easier.

\begin{Lemma}\label{l:13}
The functions $\cdot$ and $+$ are 
 associative and symmetric, and $\cdot$ distributes over $+$  on
$D^{(1)}\times D^{(1)}$.
\end{Lemma}
{\em Proof:}
Consider associativity for $+$: If  $([a,b]+[c,d])+[f,g]=[h,i]$ and 
 $[a,b]+([c,d]+[f,g])=[h',i']$, then
 $[h,i]=[e \cdot (e \cdot a \cdot d+e \cdot c \cdot b) \cdot g+e \cdot f \cdot e \cdot b \cdot d, 
e \cdot e \cdot b \cdot d \cdot g]$ and
$[h',i']=[e \cdot (e \cdot c \cdot g+e \cdot d \cdot f) \cdot b+e\cdot 
a \cdot d \cdot g \cdot e,
e \cdot b \cdot e \cdot d \cdot g]$.
These two are equal.
The other cases are similar.

Our next embedding step introduces subtraction and negative values:
 Let $D^{(2)}=(D^{(1)}\times (D^{(1)})/\approx$,
where, for
$a,b,c,d\in D^{(1)}$, $(a,b)\approx (c,d)$ iff $a+ d=b + c$. Use
notation
$[[a,b]]$ for $[(a,b)]_\approx$. An element $d\in D^{(1)} $ is identified
with
$[[d,0]]\in D^{(2)}$.
Define
$<$, $\cdot$ and $+$ in this extension by
$[[a,b]]<[[c,d]]$ iff $a + d<c + b$,
$[[a,b]]\cdot[[c,d]]=[[a \cdot c + b \cdot d,a \cdot d + b \cdot
c]]$, and
$[[a,b]]+ [[c,d]])=[[a + c,b +  d]]$.

\begin{Lemma}\label{l:emb2}
The relation $\approx$ is an equivalence relation and
if $[[d,0]]=[[d',0]]$, then $d=d'$.
\end{Lemma}
\begin{Lemma}\label{l:total}
The relation $<$ is  a total order on $D^{(2)}$ and extends $<$ on
$D^{(1)}$.
\end{Lemma}
\begin{Lemma}\label{l:14}
The relations $\cdot$ and $+$ are 
total functions on $D^{(2)} \times D^{(2)}$ and extend $\cdot$ and $+$ on 
$D^{(1)}\times D^{(1)}$.
\end{Lemma}
\begin{Lemma}\label{l:15}
The functions $\cdot$ and $+$ are 
 associative, symmetric and  $\cdot$ distributes over $+$ on 
$D^{(2)}\times D^{(2)}$.
\end{Lemma}

 \begin{Lemma}\label{l:ring}
The structure $R=(D^{(2)},<,\cdot,+,1,0)$ is an ordered integral domain.
\end{Lemma}
\begin{Lemma}(\cite[Ch V.2, Theorem 6]{Birkhoff})\label{l:ringo}
Every ordered integral domain can be embedded in an ordered field
\end{Lemma}

The structure $R$ can thus be embedded
in an ordered field. This field  embeds
the  plausibility space. 
This finishes the  proof of Theorem~1.

\section{Proof of Theorem~2}

Let the plausibility space be
$(D,\cdot,+,S,\leq,\bot,\top)$.
Most of the properties of a proper ordered plausibility space are
immediate consequences of our stated assumptions.
The non-trivial part is to show that the 
stated laws for $F$, $G$ and $S$ must hold in a consistently refinable model.
This follows from a sequence of Lemmas, each showing how
an algebraic law follows from the corresponding law of
propositional logic. We only state them and prove some of them:

 \begin{Lemma}\label{l:compl}
For all $x\in D$, $S(S(x))=x$.
\end{Lemma}
{\em Proof:}
If $x=A|B$, then $x=A|B=\overline{\overline{A}}|B=S(S(x))$.
 \begin{Lemma}\label{l:Fass}
For all $x,y,z\in D$, $(x\cdot y)\cdot z)=x\cdot (y \cdot z))$.
\end{Lemma}
{\em Proof:}
If we have worked out a model where
$ a\cdot (b\cdot c)\neq (a\cdot b)\cdot c$ for some plausibilities $a$, $b$ and $c$ that occur in the model,
then we take an arbitrary statement $S$ (not false) and refine with
$A'$, $B'$ and $C'$: $A'\rightarrow B'$,  $B'\rightarrow C'$
and   $C'\rightarrow S$,
$A'|B'=a$, $B'|C'=b$ and $C'|S=c$. Now the value $A'B'C'|S$ can be
computed in two ways giving different results, as $(A'B')C'|S=
F(A'B'|SC',C'|S)=F(F(A'|B',B'|C'),c)=(a\cdot b)\cdot c$
and as
$A'(B'C')|S=F(A'|SB'C',B'C'|S)=F(A'|B',F(B'|C',C'|S))
=a\cdot (b\cdot c)$.

 \begin{Lemma}\label{l:Fcom}
For all $x,y\in D$, $x\cdot y=y \cdot x$.
\end{Lemma}
{\em Proof:}
If $x=A|D$ and $y=B|E$, introduce by refinement
$B'$ such that $D\rightarrow B'$, $B'|D=y$ and $B'$ is independent of $A$. Now $A B'|D=x \cdot y$ and
 $A B'|D=B' A|D=y\cdot x$, so $x\cdot y=y\cdot x$.

 \begin{Lemma}\label{l:Gcom}
For all $x,y\in D$, $x+ y=y + x$,
in the sense that if one side is defined, then the other side is defined
and equal.
\end{Lemma}
{\em Proof:}
If $x=A|D$ and $y=B|E$ and $x\leq S(y)$, introduce by refinement
$B'$ such that $D\rightarrow B'$, $B'|D=y$ and $AB=\bot$. Now $A\lor B|D=x+y$ and
 $A\lor B|D=B\lor A|D=y+x$, so $x+y=y+x$.
\begin{Lemma}\label{l:Gass}
For all $x,y,z\in D$, $(x+ y) + z = x + (y+ z)$,
in the sense that if one side is defined, then the other side is defined
and equal.
\end{Lemma}
{\em Proof:}
By refinement and associativity of $\lor$.

 \begin{Lemma}\label{l:FGdistr}
For all $x,y,z\in D$, $x\cdot z + y \cdot z=(y + x)\cdot z$,
in the sense that if $x+y$ is defined, then both sides are defined
and equal.
\end{Lemma}
{\em Proof:}
Assume $x$, $y$ and $z$ are non-trivial plausibilities of the model
and $x\leq S(y)$, because 
otherwise the Lemma is obvious.
Introduce for non-$\bot$ $D$ by refinement $A\rightarrow D$, 
$B\rightarrow D $ and $C\rightarrow D $ such that $A\land B=\bot$
and $A$ and $B$ are independent of $C$, and $x=A|D$, $y=B|D$ and
$z=C|D$.
Now, $(A\lor B) \land C|D=(x+y)\cdot z$ and 
$(A\lor B) \land C|D = (A\land C)\lor (B \land C)|D=x\cdot z + y\cdot z$
and the Lemma follows.

\end{document}